\documentclass{article}

\usepackage{microtype}
\usepackage{graphicx}
\usepackage{subcaption}
\usepackage{booktabs} 

\usepackage{hyperref}

\usepackage[preprint]{icml2026}

\usepackage{amsmath}
\usepackage{amssymb}
\usepackage{mathtools}
\usepackage{amsthm}

\usepackage[capitalize,noabbrev]{cleveref}

\theoremstyle{plain}

\theoremstyle{definition}

\theoremstyle{remark}

\usepackage[disable,textsize=tiny]{todonotes}

\icmltitlerunning{Harvest: Opportunistic Peer-to-Peer GPU Caching for LLM Inference}

\begin{document}

\twocolumn[
  \icmltitle{Harvest: Opportunistic Peer-to-Peer GPU Caching for LLM Inference}



  \icmlsetsymbol{equal}{*}

  \begin{icmlauthorlist}
    \icmlauthor{Nikhil Gopal}{columbia}
    \icmlauthor{Kostis Kaffes}{columbia}
  \end{icmlauthorlist}

  \icmlaffiliation{columbia}{Department of Computer Science, Columbia University, New York, United States}

  \icmlcorrespondingauthor{Nikhil Gopal}{nsg2127@columbia.edu}
  \icmlcorrespondingauthor{Kostis Kaffes}{kkafes@cs.columbia.edu}

  \icmlkeywords{Machine Learning, ICML, LLM, Transformers, KV cache Offload}

  \vskip 0.3in
]



\printAffiliationsAndNotice{}  

\begin{abstract}
  
  Large Language Model (LLM) inference is increasingly constrained by GPU memory capacity rather than compute throughput, driven by growing model sizes and the linear growth of the key–value (KV) cache during autoregressive decoding. Existing approaches mitigate memory pressure by offloading model state and KV tensors to host memory, but incur substantial latency due to limited PCIe bandwidth. We present \emph{\textbf{Harvest}}, an opportunistic GPU cache management framework that exploits high-bandwidth peer-to-peer GPU interconnects to dynamically place model weights and KV cache in unused GPU memory. Harvest treats peer GPU memory as a transient cache tier, preserving correctness while reducing data movement overhead under dynamic memory availability. We demonstrate significant throughput speedup of more than $2\times$ by using Harvest to accelerate the retrieval of two widely-used inference components: expert layer weights and KV cache entries.
  
\end{abstract}

\section{Introduction}

Memory capacity increasingly limits GPU-based serving of large language models (LLMs).
Each GPU must store model weights, per-request runtime metadata, and the transformer key--value (KV) cache used by attention during decoding.
The KV cache grows with context length, and the combined footprint can exceed the high-bandwidth memory (HBM) capacity of a single GPU for large models or long contexts, even when arithmetic throughput remains available.
When HBM becomes the bottleneck, systems reduce batch size, shorten context, or accept higher latency.

Existing systems recover capacity by moving data off GPU memory. Many deployments offload weights or the KV cache to host DRAM and fetch data over PCIe on demand, which preserves correctness but places PCIe bandwidth and latency on the critical path. Other deployments shard the model across additional GPUs, which raises hardware cost and increases synchronization and communication overhead. These designs therefore exchange memory capacity for latency or cost.

This paper presents \emph{Harvest}, a cache management framework that uses remote GPU memory as an intermediate tier for memory-heavy inference state.
Harvest treats unused HBM on a peer GPU (another GPU in the same server or rack, connected to the compute GPU via NVLink) as a best-effort cache.
Harvest transfers cached objects over NVLink, a high-bandwidth GPU-to-GPU interconnect.
The Harvest runtime monitors peer memory availability, places objects in peer HBM when capacity exists, and revokes allocations when capacity disappears.
Correctness never depends on the peer tier: an application either (i) retains an authoritative copy in host DRAM or (ii) tolerates data loss and reconstructs the object after revocation.
Harvest therefore targets multi-GPU servers and NVLink fabrics where free memory on some GPUs can coexist with memory pressure on others.

\begin{figure}[h]
    \centering
    \includegraphics[width=0.75\linewidth]{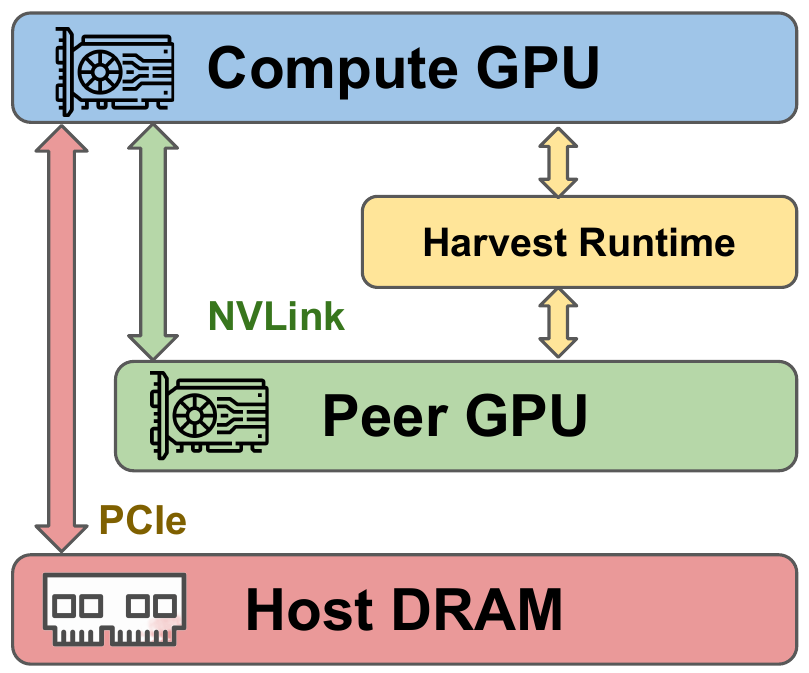}
    \caption{Harvest enables the use of fast remote NVLink-connected GPU memory as an opportunistic cache.}
    \label{fig:harvest-intro}
\end{figure}

We evaluate Harvest on two representative LLM inference workloads: (i) KV cache offloading for long-context decoding and (ii) expert offloading for mixture-of-experts (MoE) models.
Harvest improves MoE expert transfer latency by up to almost $10\times$ and increases end-to-end throughput by $1.5$--$2.0\times$ for models such as Qwen and Phi-3.5.
Harvest also reduces KV cache transfer latency by up to $5.65\times$ vs. CPU offload.

In summary, this paper makes three contributions: (1) a placement model that exposes peer GPU memory as a transient cache tier, (2) a runtime system that adapts placement to dynamic availability without requiring model code changes, and (3) an empirical evaluation of peer-memory caching for LLM inference.

\section{Motivation and Related Work}
\label{sec:motivation}

Compared to host-based offloading over PCIe, peer GPU access over NVLink significantly reduces the latency of servicing page faults and fetching model state, making it a natural substrate for latency-sensitive inference workloads. Prior work has shown that faster interconnects can materially improve inference performance when memory movement dominates execution~\cite{fang2025acceleratingllminferencedynamic, kumar2025aqua}.

Peer-GPU caching helps when two conditions are met.
A deployment must contain spare GPU memory that can absorb overflow from a memory-bound GPU.
The deployment must also provide an interconnect that moves cached state fast enough to stay off the critical path.

\begin{figure}[t]
    \centering
    \includegraphics[width=1.0\linewidth]{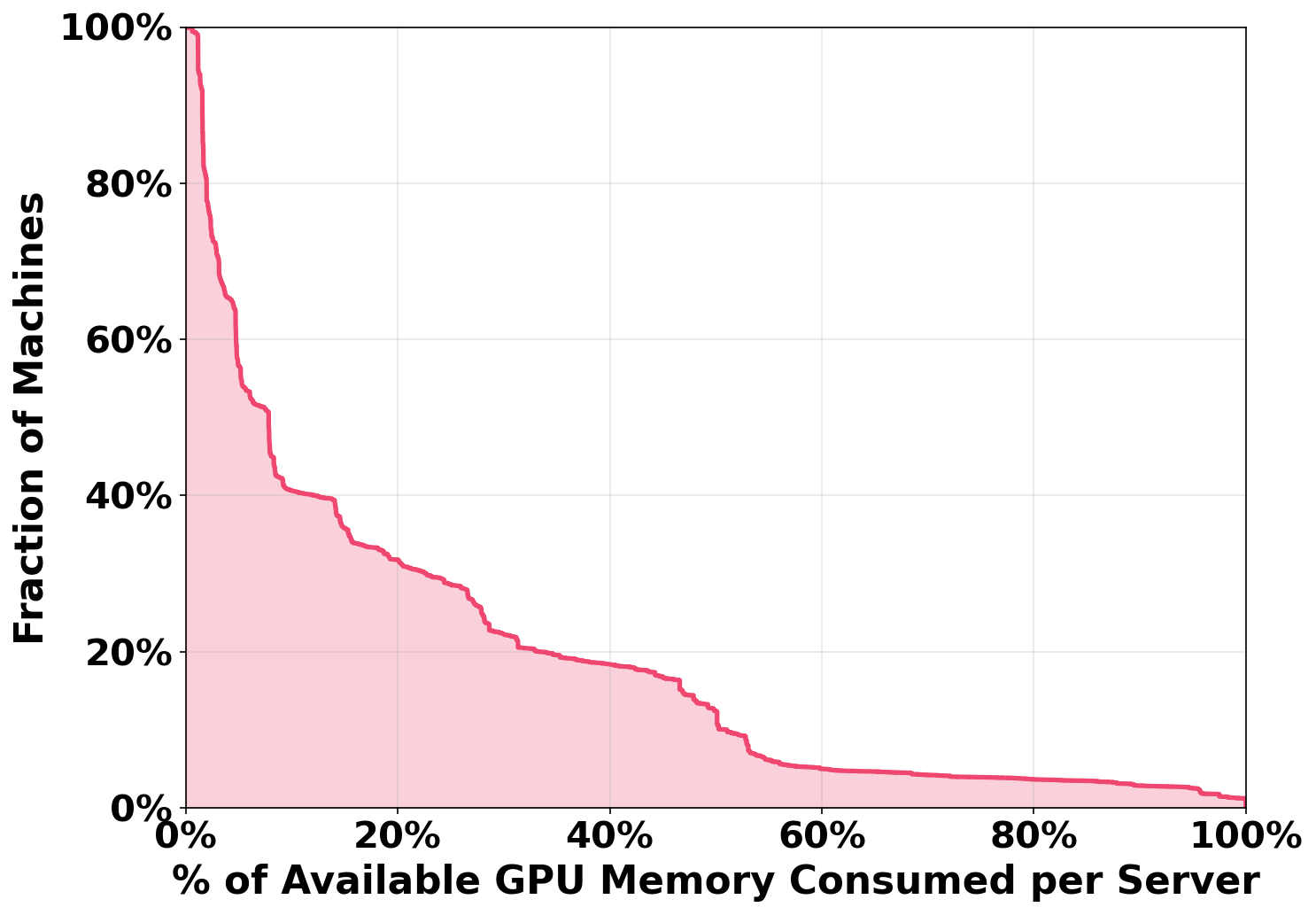}
    \caption{CDF of GPU Memory Consumption across Alibaba Cloud Training and Inference Cluster.}
    \label{fig:avail-gpumem}
\end{figure}

\subsection{Production traces show spare GPU memory}
\label{sec:motivation:memory}
Cloud GPU clusters multiplex heterogeneous jobs with heterogeneous memory footprints.
The \texttt{gpu-v2020} dataset in the \href{https://github.com/alibaba/clusterdata/tree/master}{Alibaba Cluster Trace Program} logs GPU memory usage across 6{,}500 GPUs on 1{,}800 machines running a mix of training and inference workloads~\cite{weng2022mlaas}.
Figure~\ref{fig:avail-gpumem} summarizes 959{,}080 machine snapshots and plots the fraction of machines that exceed a given GPU-memory-consumption level.

Figure~\ref{fig:avail-gpumem} shows that in many machines applications consume a small fraction of available GPU memory.
About 68\% of the machines consume at most 20\% of the available GPU memory, and about 87\% of machines consume at most 50\% of the GPU memory.

The \texttt{gpu-v2020} trace captures a 2020-era workload mix, and newer measurements report similar dispersion in GPU memory footprints.
FlexPipe analyzes two weeks of GPU resources from a major cloud provider and reports mean GPU memory utilization of 43.48\% (median 28.78\%) in an inference-only cluster, with 38.44\% of samples in the 10--30\% utilization bin~\cite{flexpipe2026}.
MuxFlow reports production deployment at ByteDance and reports GPU memory usage of 42\% (increasing to 48\% after enabling space sharing), which implies substantial headroom even in a large production cluster~\cite{muxflow2024}.
Other environments can run closer to capacity; an NSDI'24 study of LLM development in the datacenter reports that 50\% of GPUs in Kalos consume over 75\% of GPU memory (60\,GB)~\cite{hu2024llmdev}.

Together, these newer sources show that GPU memory headroom and heterogeneity persist in modern deployments, while the amount of spare memory depends on workload mix and cluster policy.
Diurnal and seasonal demand patterns shift the mix of active jobs, and multi-tenant schedulers pack jobs with different memory demands onto the same cluster~\cite{306025}.
Varying load directed to LLM inference servers also leads to lower memory usage when the KV cache utilization is low~\cite{kumar2025aqua}.
A cluster therefore contains memory-heavy GPUs and memory-light GPUs at the same time, which creates opportunities to borrow memory without changing the hardware footprint.

\begin{figure}[t]
    \centering
    \includegraphics[width=1\linewidth]{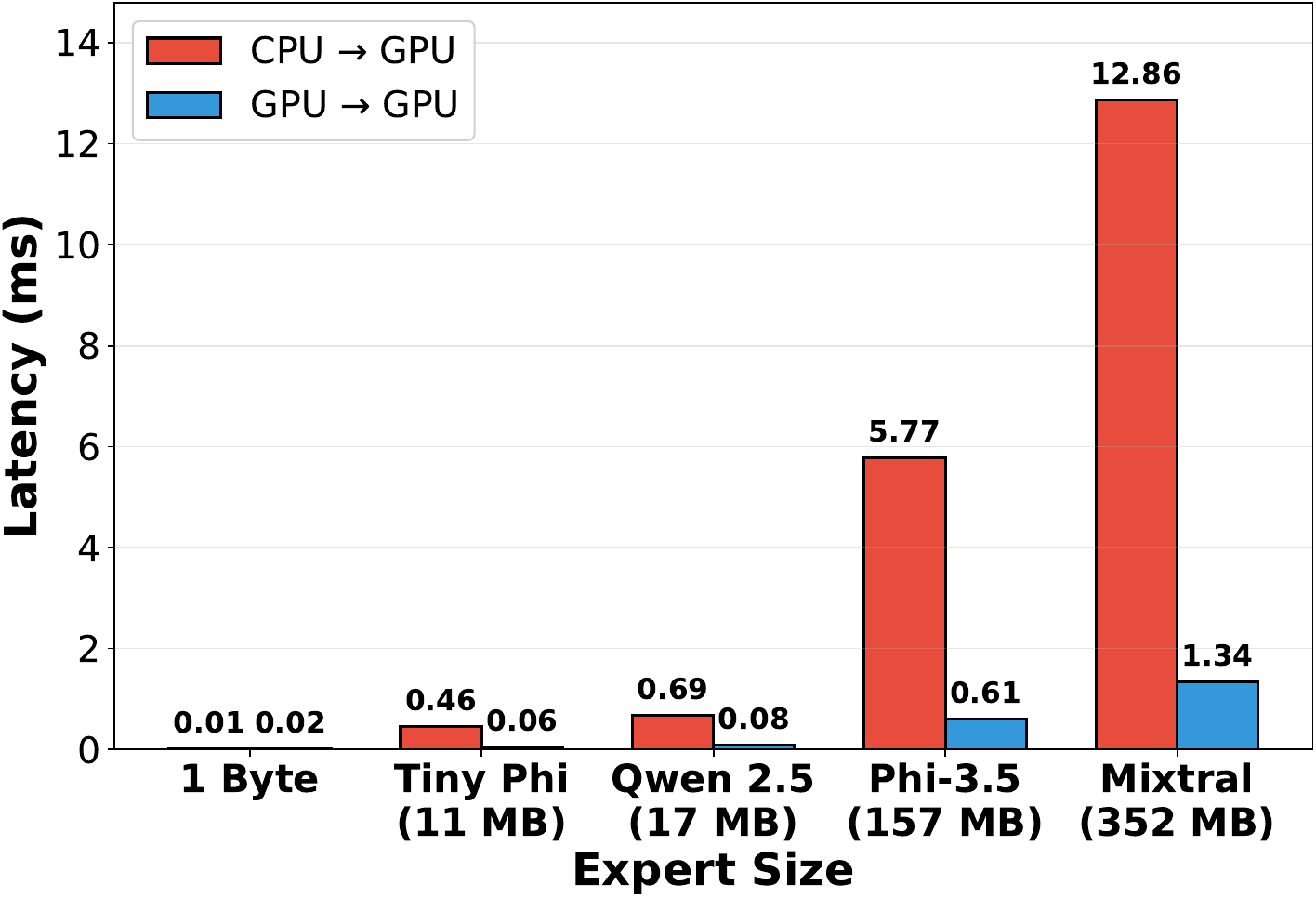}
    \caption{GPU$\leftrightarrow$GPU and GPU$\leftrightarrow$CPU transfer latency of memory chunks of different sizes, mapped to expert sizes of different MoE models for reference.}
    \label{fig:moe_latency}
\end{figure}

\subsection{NVLink makes peer GPU memory a fast tier}
\label{sec:motivation:speed}
Host offloading uses PCIe to fetch state from CPU-attached DRAM.
PCIe transfers add latency and contend with other traffic because the PCIe link sits on the critical path of a cache miss.
NVLink changes the cost model by enabling direct GPU-to-GPU transfers over a peer-to-peer interconnect that provides higher bandwidth and lower latency than PCIe for on-node transfers.

Figure~\ref{fig:moe_latency} shows the latency of moving memory chunks of various sizes, mapped to expert sizes of different MoE models for reference.
The speed-up is consistently high, ranging from 7.5$\times$ from the very small Tiny Phi model~\cite{abdin2024phi3technicalreporthighly} to 9.5$\times$ for the much bigger Mixtral 8$\times$7B model~\cite{jiang2024mixtralexperts}.

All testing for this paper was conducted using a Microsoft Azure NC80adis H100 v5 Virtual Machine with 80 vCPUs, 640 GB of RAM and two Nvidia H100 GPUs. The machines used PCIe version 5.0 with 12 NVLink links between the 2 GPUs. 

Together, Figure~\ref{fig:avail-gpumem} and Figure~\ref{fig:moe_latency} motivate a cache tier built from remote GPU memory.
Spare memory appears frequently in production traces, and NVLink supplies a fast path to access that memory.
Harvest exploits this combination by treating peer GPU memory as a transient cache that reduces PCIe traffic while preserving correctness when peer memory availability changes.

Future deployments will increase the size of the NVLink domain and therefore the cache capacity that we can harvest.
Larger NVLink domains increase the pool of peer HBM and increase the probability that at least one peer GPU has spare capacity at a given time.
NVIDIA already ships rack-scale designs that connect dozens of GPUs with NVLink and NVSwitch, including systems that connect 72 GPUs in an all-to-all topology~\cite{nvidia_gb200_nvl72}.
NVIDIA also describes NVLink Switch Systems that interconnect up to 256 GPUs~\cite{nvidia_nvlink_switch_system}.

\subsection{Related Work}
\label{sec:motivation:related}

\paragraph{\textbf{LLM memory management and offloading.}}
PagedAttention in vLLM reduces KV cache waste by paging KV into fixed-size blocks and enabling sharing across requests~\cite{10.1145/3600006.3613165}. FlexGen and other host-offload engines aggregate GPU, CPU, and disk memory to run models that exceed HBM capacity, then stream tensors over PCIe during execution~\cite{sheng2023flexgenhighthroughputgenerativeinference}. KVPR reduces PCIe stall time by overlapping transfers with partial KV recomputation on the GPU~\cite{jiang-etal-2025-kvpr}.
Aqua pushes the offload tier into the scale-up GPU domain by moving the KV cache into the HBM of other GPUs connected via NVLink/NVSwitch, and it accelerates paging to support preemptive, time-sliced inference~\cite{kumar2025aqua}. CUDA Unified Memory also enables oversubscription through driver-managed page migration in a unified virtual address space, but page faults and migration introduce variable latency that can land on the critical path~\cite{cuda_unified_memory}.

\paragraph{\textbf{MoE serving systems.}}
MoE-Lightning improves MoE batch inference on memory-limited GPUs with a CPU--GPU--I/O pipelining schedule (CGOPipe) and policy selection via a performance model (HRM)~\cite{10.1145/3669940.3707267}. MoE-Infinity traces sparse expert activation and uses expert caching and prefetching to reduce host$\rightarrow$GPU traffic during decoding~\cite{xue2025moeinfinityefficientmoeinference}. Compression recipes further reduce MoE footprint through expert trimming and more aggressive layer/block dropping~\cite{he2025efficientmixtureexpertsholistic}.

\section{Harvest Design}
\label{sec:harvest-design}

\subsection{General Harvesting Model}
\label{sec:harvest-model}

Harvest provides an abstraction for opportunistic GPU memory harvesting.
Harvest exposes unused high-bandwidth memory (HBM) on \emph{peer GPUs} to a user-space application as a best-effort cache tier.
The design assumes three memory tiers: compute GPU HBM (authoritative), peer GPU HBM (ephemeral cache), and host DRAM (optional backing store).
Correctness never depends on the peer tier.
Peer allocations may disappear at any time, so every cached object must either be reconstructible or backed by an authoritative copy elsewhere.

User-space applications interact with Harvest through a Harvest API that exposes an opportunistic allocation primitive and a revocation notification mechanism.
An application calls Harvest to request space in peer HBM, and a controller selects a peer device and performs the allocation.
On success, Harvest returns a handle that identifies the peer device, the device pointer, the allocation size, and metadata needed to locate the object.
The application may register a revocation callback for each allocation.
The runtime triggers the callback when the peer allocation becomes unavailable, which allows the application to remove the mapping and fall back to an authoritative source.

\paragraph{Consistency is an application choice.}
Harvest does not maintain coherence or consistency for peer-resident data.
The runtime never tracks dirty state and never performs write-back.
The application selects one of two outcomes when a cached object is revoked: (i) serve the object from an authoritative copy in host DRAM, or (ii) lose the object and reconstruct it later.
Both outcomes occur in LLM inference.
Lossy caching matches transient state such as intermediate activations or KV cache entries that can be recomputed or dropped without correctness violations, while backed caching matches large, expensive-to-reconstruct state such as weights or MoE expert parameters.

\begin{figure}[t]
    \centering
    \includegraphics[width=1.0\linewidth]{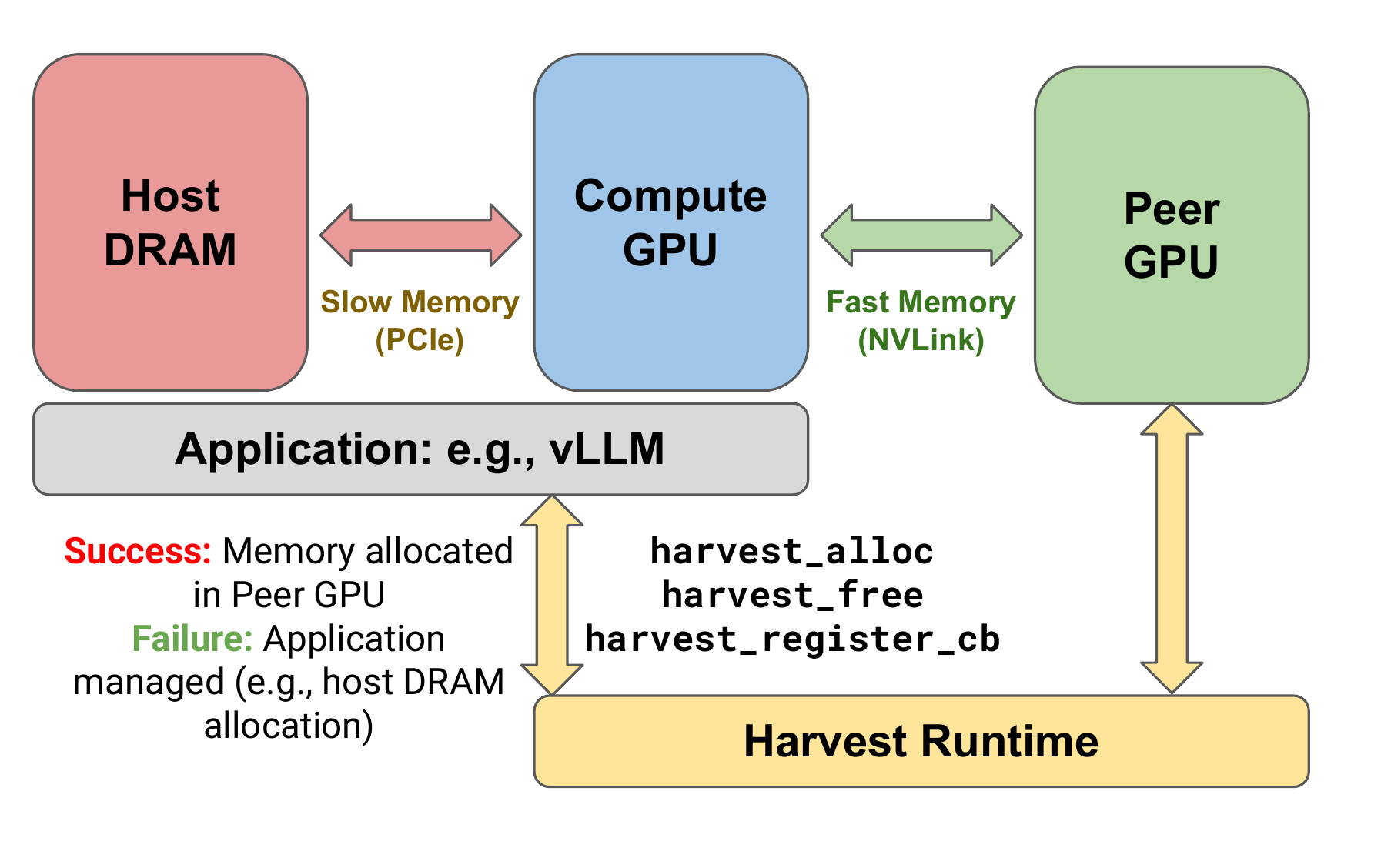}
    \caption{Harvest architecture. The Harvest runtime spans the compute and peer GPUs under the same NVLink domain and responds to application-initiated memory allocation requests by harvesting available fast memory in a peer GPU. If the allocation fails or is revoked, the application can revert to the slow host DRAM.}
    \label{fig:harvest}
\end{figure}

\subsection{Harvest API and Runtime Workflow}
\label{sec:harvest-api}

The Harvest API exposes two core operations: an opportunistic remote allocation and a revocation registration mechanism.

\begin{center}
\begin{tabular}{ll}
\texttt{harvest\_alloc(size, hints)} & \\
\texttt{harvest\_free(handle)} & \\
\texttt{harvest\_register\_cb(handle, cb)} &\\
\end{tabular}
\end{center}

Internally, \texttt{harvest\_alloc} selects a peer GPU with harvestable capacity, switches to that device context, and allocates memory using the standard CUDA allocation path.
Harvest returns a \texttt{(device, pointer, size)} tuple plus metadata that the application records in a placement map.
The tuple constitutes a unique \texttt{handle} that is used for managing deallocation. 
All data movement remains explicit.
The application moves data with CUDA peer-to-peer copies (e.g., \texttt{cudaMemcpyPeerAsync}).
Harvest never requires kernels on one GPU to dereference pointers that reside on another GPU.

Revocation and deallocation are explicit and ordered.
A peer allocation may be freed due to allocator pressure, policy-driven eviction, or external reclamation by a higher-priority workload.
Before freeing memory, the runtime drains in-flight DMA and kernel operations that touch the region (for example, by synchronizing the relevant streams or events).
After the runtime invalidates the placement entry, the runtime invokes the registered callback.
The callback updates application state and initiates fallback to host DRAM or reconstruction, depending on the object’s durability mode.

\paragraph{\textbf{Allocation policy.}}
A controller decides where and how to allocate peer memory.
The current prototype uses a best-fit strategy that chooses a peer GPU and a free segment that minimize leftover fragmentation while satisfying size and hint constraints.
The Harvest API does not require best-fit.
Other policies can optimize locality (prefer NVLink-adjacent peers), fairness (rate-limit individual clients), interference (avoid peers with high memory-bandwidth demand), or stability (prefer peers with low churn).

\paragraph{\textbf{Isolation with MIG.}}
Harvest isolates harvested capacity from co-located workloads on the peer GPU using NVIDIA Multi-Instance GPU (MIG).
MIG partitions a physical GPU into hardware-isolated instances with dedicated memory-system paths and dedicated compute resources, which limits cross-tenant interference and provides fault isolation~\cite{nvidia_mig_guide_intro}.
Harvest reserves one MIG instance as a cache device and allocates harvested memory only inside that instance.
Other workloads run in separate instances, so cache allocations cannot consume their HBM budget or thrash their cache and memory bandwidth allocation.
Some driver configurations restrict peer-to-peer communication for MIG devices (for example, cross-GPU P2P may be unavailable), so Harvest treats MIG as a deployment choice for isolation rather than a functional requirement for harvesting~\cite{nvidia_mig_deploy}.

\section{Harvest for MoE Offload}

\subsection{Mixture-Of-Experts Architecture}

Mixture-of-Experts (MoE) has emerged as a popular variant of the Transformer architecture that enables parameter scaling without a proportional increase in per-token computation. Instead of using a single dense feed-forward network, MoE layers consist of a set of $N$ expert networks, of which only a small subset is activated for each token~\cite{shazeer2017}. A learned gating function routes each input token to its top-k experts, typically with $k \ll N$, allowing the model to increase total parameter count while keeping the computational cost per token approximately constant. As a result, MoE architectures have been widely adopted by frontier labs such as Mistral AI and DeepSeek and are among the most popular language models used today~\cite{10937907}.

\subsection{MoE's Large Memory Footprint}

Driven by scaling laws that correlate parameter count with model performance, Mixture of Experts (MoE) architectures are growing rapidly in size~\cite{clark2022unifiedscalinglawsrouted}. However, the deployment of modern MoEs can quickly exhaust GPU memory, even with modern hardware. For example,~\citet{yan2025acceleratingmixtureofexpertinferenceadaptive} note that deploying DeepSeek-R1, with its 671B parameters, requires 16 NVIDIA A100 (80GB) GPUs solely due to its memory footprint. Newer MoE models such as Kimi-K2 have surpassed 1 Trillion parameters, ensuring that memory pressure at inference time will continue to increase~\cite{kimiteam2025kimik2openagentic}.


Expert access patterns are highly skewed and exhibit temporal locality: certain experts are frequently activated, while others remain unused~\cite{shazeer2017}. Crucially, this skew is dynamic.~\cite{doucet2025harmoenyefficientmultigpuinference} observe that token distributions vary significantly across queries, causing expert ``hotspots'' to shift unpredictably.~\citet{wu2026acceleratingedgeinferencedistributed} further demonstrate that in heterogeneous serving environments, expert activation patterns diverge based on specific user tasks (e.g., arithmetic vs. code generation), causing static placement policies to degrade performance when workloads drift.

Prior work has leveraged this sparsity to enable inference on memory-constrained hardware by offloading experts to host DRAM. For instance, ~\citet{yan2025acceleratingmixtureofexpertinferenceadaptive} propose MoEpic, a system that segments experts into cached ``top'' and offloaded ``bottom'' halves to optimize hit rates under limited VRAM. However, the imbalanced token routing means offloading to DRAM quickly saturates PCIe bandwidth, causing the GPU to starve~\cite{liu2025bandwidthefficientadaptivemixtureofexpertslowrank}. 

The combination of high memory consumption and sparse, dynamic activations underscores the motivation for Harvest. By replacing slow PCIe fetches with opportunistic, high-bandwidth peer-to-peer transfers, we can reduce the latency penalty of fetching non-resident experts while reducing the memory pressure of the ever increasing LLM memory footprints.

\subsection{MoE Offloading Design}
\label{sec:expert-rebalance}

MoE Lightning is an inference framework for Mixture-of-Experts models, and CGOPipe is its pipelined execution strategy for overlapping offloaded expert weight fetches with compute~\cite{10.1145/3669940.3707267}. This is accomplished by partitioning batches into micro-batches, and subsequently overlapping expert weight transfers for micro-batch $i{+}1$ with GPU computation for micro-batch $i$. Expert weights are paged at expert granularity, and an entire expert’s parameters must be resident in GPU memory before its feedforward computation can execute. Harvest does not modify CGOPipe’s routing, batching, CPU-side attention computation, or pipeline structure. Rather, Harvest extends CGOPipe by adding peer GPUs as a device for offloaded expert weights, serving expert cache misses from peer GPU memory instead of host DRAM when available.

The Expert Rebalancer applies the Harvest API to Mixture-of-Experts (MoE) model weights.
At server start, a user-defined subset of experts is loaded into local HBM, while the remaining experts reside in host DRAM.
As peer memory becomes available, the rebalancer allocates peer GPU memory using \texttt{harvest\_alloc} and migrates selected expert weights into peer HBM.

Expert placement is tracked using an expert residency map that records, for each expert, whether it resides in local HBM, peer HBM, or host DRAM. Upon a cache miss, the runtime does not automatically fetch expert weights to peer HBM. If a peer allocation is revoked, the rebalancer invalidates the corresponding residency entry, and future invocations automatically fall back to pinned host DRAM.

\subsection{MoE Offloading Evaluation Setup}
All of our benchmarks were computed using MoE Lightning's test bench with \texttt{--max-new-tokens=32}, a micro-batch size of $\mu = 324$ tokens and $b = 14$ micro-batches, resulting in a total batch size of $N = \mu \times b = 4{,}536$ tokens. Prompts were drawn from the MTBench dataset, matching MoE-Lightning's evaluation setup~\cite{Bai_2024}. Results were computed as an average of the throughput across 5 trials of the test bench, each of which included generating 50 tokens as a "warmup" to account for cold starts.

We tested Mistral AI's Mixtral-8x7B-Instruct-v0.1, Alibaba's Qwen2-MoE, and Microsoft's Phi-3.5-MoE-instruct and 
Phi-tiny-MoE-instruct~\cite{jiang2024mixtralexperts, yang2024qwen2technicalreport, abdin2024phi3technicalreporthighly}.
No modifications were made to the expert routing mechanisms present in the base models.
Though some of the smaller models could fit entirely into a single H100's HBM, we forced the testing configuration to cache expert weights in the compute GPU. This was done to force expert offloading for the smaller models, and to measure how the effects of expert offloading scale with memory consumption. We evaluate on the hardware setup described in Section ~\ref{sec:motivation:speed}. 

\begin{table}
\caption{MoE Model Architecture Comparison}
\label{tab:moe-models}
\centering
\resizebox{\columnwidth}{!}{%
\begin{tabular}{lcccc}
\toprule
Model & Params (B) & Active (B) & Experts & Active Exp. \\
\midrule
\textbf{Mixtral-8x7B}   & 47.0 & 13.0 & 8  & 2 \\
\textbf{Phi-3.5-MoE}   & 60.8 & 6.6  & 16 & 2 \\
\textbf{Phi-tiny-MoE}  & 3.8  & 1.1  & 16 & 2 \\
\textbf{Qwen2-MoE}     & 14.3 & 2.7  & 64 & 4 \\
\bottomrule
\end{tabular}}
\end{table}

\subsection{Expert Offloading Results}

\begin{figure}
    \centering
    \includegraphics[width=1\linewidth]{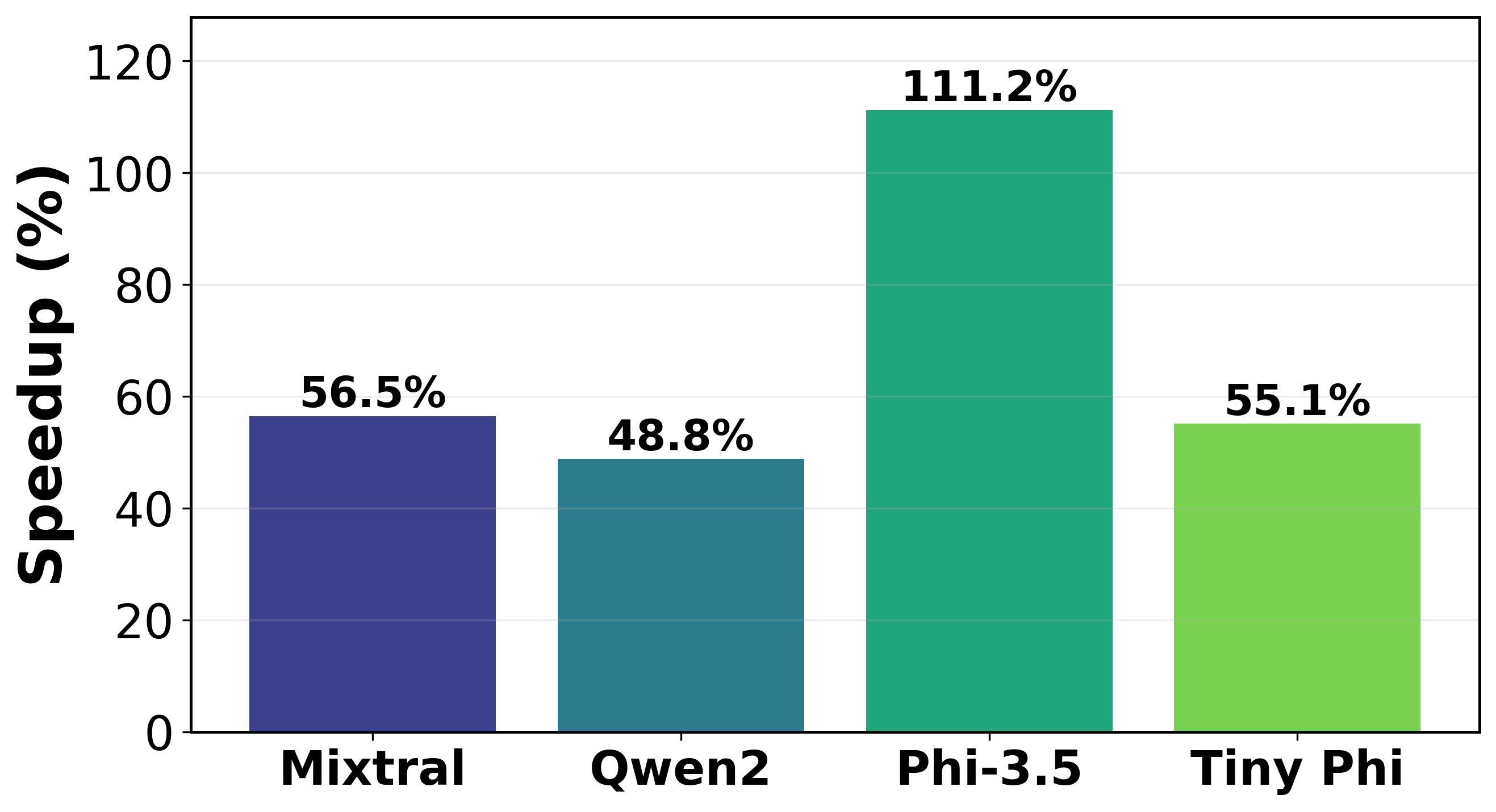}
    \caption{Token generation throughput improvement with expert weight offloading to peer GPU using Harvest vs to CPU using CGOPipe. In this scenario, we force 50\% experts to be offloaded.}
    \label{fig:expert_offload_speedup}
\end{figure}

Figure~\ref{fig:expert_offload_speedup} shows that extending CGOPipe with peer GPU expert caching yields substantial decode throughput improvements across all evaluated MoE models, ranging from 48\% to over 110\%. These gains are achieved without modifying routing, batching, or attention kernels, isolating the impact to peer offloading alone. These numbers reflect averaged results across 5 runs of the MoE Lightning test bench with half of the experts offloaded.

Notably, Phi-3.5-MoE exhibits nearly double the speedup of Qwen2-MoE despite having comparable parameter count. This disparity is explained by architectural differences rather than parameter count alone: Phi-3.5-MoE has fewer experts and a smaller top-\emph{k} routing fan-out, resulting in higher temporal locality and more frequent reuse of expert weights across micro-batches. In contrast, Qwen2-MoE activates a larger number of distinct experts per token, increasing expert working-set churn.

These results indicate that Harvest is most effective when decode latency is dominated by expert weight fetches. By replacing PCIe-based transfers with higher-bandwidth NVLink peer-to-peer transfers, Harvest alleviates this bottleneck and shifts decode execution closer to the compute-bound regime. Models with higher expert reuse thus benefit disproportionately, while models with larger expert working sets observe smaller, though still significant, gains.

\begin{figure*}[t!]
    \centering
    \includegraphics[width=\linewidth]{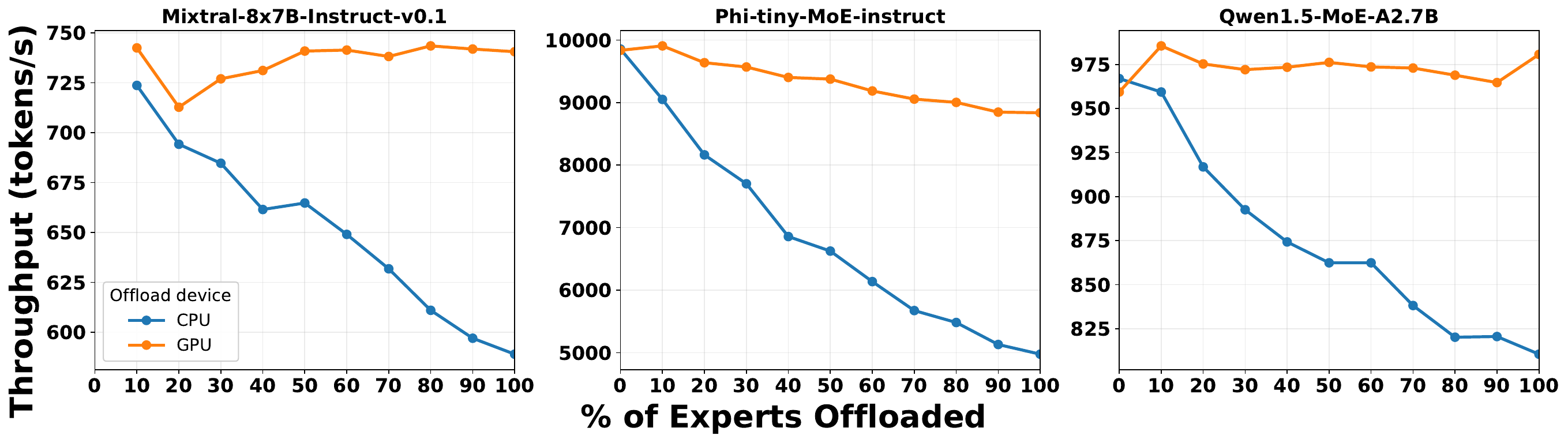}
    \caption{Throughput as a function of expert offload percentage for three representative MoE models with GPU and CPU offloading. Results for Phi-3.5-MoE are similar to Qwen1.5 and omitted for brevity.}
    \label{fig:offload-ablations}
\end{figure*}

Figure~\ref{fig:offload-ablations} further examines how throughput varies with the fraction of experts offloaded. Across all evaluated models, throughput degrades significantly with a larger offloaded fraction for CPU offloading but remains stable or degrades minimally for Harvest’s GPU offloading. For example, Qwen2-MoE's throughput remains nearly constant at approximately 975 tokens/s from 0\% to 100\% experts offloaded, whereas CPU offloading drops to about 810 tokens/s at full offload. Similarly, Mixtral maintains roughly 740 tokens/s with GPU offloading but falls below 600 tokens/s when all experts are served from host memory.

\section{Harvest for KV Cache Offload}

\subsection{KV Cache Growth Makes Decode Memory Bound}
During autoregressive decoding, each generated token appends new key/value (KV) vectors for every transformer layer. As a result, the KV cache footprint grows with sequence length (and batch size), and it can become the dominant contributor to inference-time memory for long-context workloads. Recent analyses show that for long sequence lengths, KV cache memory can overtake model weights as the primary memory bottleneck, especially under high concurrency~\cite{kang2024gear}. This pressure forces serving systems to either reduce concurrency or intelligently compress KV state~\cite{kampeas2026jointencodingkvcacheblocks}. In extreme cases, it can be more efficient to recompute the KV cache instead of fetching it from the slow path after offloading~\cite{jiang-etal-2025-kvpr}.

In a multi-GPU server rack with hardware interconnects, decoding will naturally consume memory on multiple GPUs simultaneously. This creates an opportunity to insert a \emph{best-effort} intermediate tier: keep the authoritative copy of spilled KV in host DRAM, but opportunistically place recently-evicted KV blocks in peer HBM so that misses are serviced by GPU$\leftrightarrow$GPU transfers rather than PCIe.

\subsection{KV Offloader Design}

Harvest instantiates its KV offloading abstractions by extending vLLM’s existing KV cache management and decode execution paths. We introduce a \texttt{KVOffloadManager} into vLLM’s KV manager, which serves as a pluggable control interface for implementing Harvest’s policy-driven allocation, migration, and revocation semantics. This manager exposes hooks that allow user-defined policies to decide when KV blocks should be offloaded, reloaded, or evicted in response to memory pressure and access patterns. For each device, Harvest extends vLLM with an \texttt{OffloadingHandler} responsible for executing data movement operations. Additionally, we augment vLLM’s KV metadata with a unified KV block table that maps logical block identifiers to their current residency across local HBM, peer GPU memory, or host DRAM.

During inference, decode workers consult an extended KV block table to resolve the physical location of each required KV block. If a block is not resident in local HBM, the worker issues a reload request to the corresponding offloading handler, which asynchronously transfers the block from a peer GPU using CUDA P2P copies. Decoding resumes once the handler signals completion. Under sustained memory pressure, workers similarly request block evictions, allowing handlers to migrate blocks out of local HBM. 

If a peer-resident block is revoked, the registered revocation callback invalidates the corresponding table entry, triggering a fallback to host DRAM or recomputation when more efficient. Unlike MoE weights, KV state does not require a persistent duplicate across tiers and may be materialized in host memory only upon eviction when durability is required.

\subsection{KV Offloader Evaluation}
We measure the time it takes to transfer KV cache entries of three different models (DeepSeekV3~\cite{deepseekai2025deepseekv3technicalreport}, Mistral-Large-3-675B-Base-2512~\cite{mistral-large-3} and Kimi-K2-Instruct-0905~\cite{kimiteam2025kimik2openagentic}) using local CPU (as used by vanilla vLLM) and remote GPU memory (as used by Harvest).
We assume FP16 KV precision and evaluate chunks 100, 500, 1000, 2000, 4000, and 8000 KV cache entries. For each footprint, we measure transfer latency for (i) host-to-GPU copies (CPU$\rightarrow$GPU) and (ii) peer-GPU copies (GPU$\rightarrow$GPU).

\subsection{KV Offloader Results}

\begin{figure*}[t!]
    \centering
    \includegraphics[width=\linewidth]{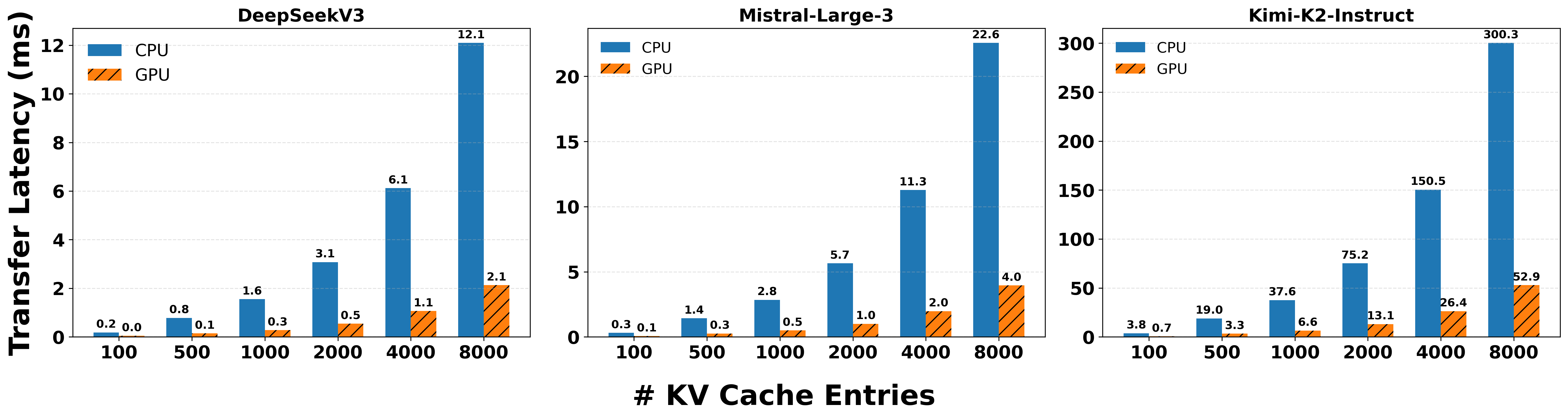}
    \caption{KV cache transfer latency for CPU vs. GPU reloads.}
    \label{fig:kv-offload-combined}
\end{figure*}

Figure~\ref{fig:kv-offload-combined} shows that the cost of moving KV-sized state grows rapidly with the number of KV cache entries, and that servicing reloads via a harvested peer GPU is consistently faster than via host memory. The gap widens as sequences grow, indicating that host-based paging over PCIe quickly becomes a dominant contributor to end-to-end decode latency for long-context workloads. On the Kimi-K2 model, the speedup ranges from approximately $5.42\times$ at 100 KV entries to $5.68\times$ at 8000 KV entries. For Mistral-Large-3, the speedup ranges from approximately $3\times$ to $5.65\times$ over the same number of KV entries.

These results support Harvest’s design choice to treat peer GPU memory as an opportunistic cache tier. When available, recently-evicted KV blocks can be placed in peer HBM and reloaded over NVLink, reducing both the frequency and cost of cache misses. 

\section{Discussion}
\subsection{Opportunistic Peer Memory as a Cache Tier}
Our results suggest that the main value of peer-to-peer GPU memory in inference is not merely higher bandwidth, but the ability to introduce a \emph{best-effort} intermediate tier that reduces the cost of evictions and page faults without changing correctness semantics.
Harvest leverages a property shared by both workloads we study: \textbf{large inference-time state is \emph{performance critical but non-authoritative}}.
This enables opportunistic placement decisions that improve throughput when remote HBM is available, while preserving correctness under eviction and volatility.

This design differs from conventional multi-GPU execution models where remote GPU state is correctness-critical and tightly coupled with routing or synchronization. By decoupling correctness from remote placement, Harvest can exploit spare capacity when present, while falling back to established CPU-backed paging mechanisms when it disappears. This perspective suggests a broader design principle for LLM inference systems: when large state is reconstructible or safely evictable, exposing it to a volatile, high-bandwidth cache tier can improve utilization without introducing new failure modes.

\subsection{\textbf{When to Harvest}}
\textbf{Harvest is most effective in regimes with (i) \emph{high reuse} of evicted state and (ii) \emph{high eviction pressure}.} Both conditions arise naturally in multi-tenant or multi-process serving scenarios: shared prompt prefixes induce repeated access to the same KV pages, while tight memory budgets and high concurrency force frequent eviction. Under these conditions, servicing faults from peer HBM rather than host DRAM reduces the dominant latency component of decode. In contrast, workloads with little temporal locality (e.g., unique prefixes) see smaller gains because evicted state is rarely reused; in these cases, additional caching capacity provides limited benefit regardless of the tier.

\subsection{Completely Fair Decoding}
Introducing token-level preemption increases fairness but also increases interleaving between active sequences, which can amplify churn in the KV working set. Our results indicate that this churn can move the system into a regime where fault service dominates execution, making memory-tier placement a first-order throughput determinant. In this sense, peer-HBM offloading can be viewed as a \emph{scheduler robustness mechanism}: it reduces the performance penalty of fairness-oriented scheduling by lowering the marginal cost of preemption-induced reloads. A practical implication is that systems may be able to employ finer-grained fairness policies without incurring the full throughput penalty typically associated with increased paging.

\section{Limitations}

Harvest was evaluated on a single two-GPU NVLink-connected system. We do not evaluate performance on larger NVLink fabrics (e.g., NVSwitch) where contention, routing, and synchronization overheads may alter the effective benefit of peer caching. We also do not study scenarios with significant NVLink congestion from concurrent model-parallel collectives or other tenants, which could reduce the  bandwidth available for paging. Finally, our evaluation focuses on two representative workloads, additional workloads and broader ablations over cache size, page management policy and scheduling parameters would be valuable to further generalize our results.

\section{Future Work}

The CUDA compiler provides many optimizations upon the initialization of the inference server. A fixed amount of memory is allocated and the GPU memory usage is not meant to change. By enabling dynamic offloading, we lose the benefits of these compiler optimizations such as CUDA graphs. vLLM is written in PyTorch, but it would be interesting to explore if the benefits of Just-In-Time compilation provided by a Deep Learning Framework like Google's JAX could increase performance by generating new kernels when GPU memory usage changes dynamically.

We also did not attempt to study the optimal page replacement policy for the KV cache, as it is likely workload dependent. It would be interesting to use memory accesses from profiling data to determine optimal page replacement policies, and to develop a sliding window-like algorithm that monitors a system's performance and hot-swaps policies.

Harvest is currently neither topology-aware nor aware of other workloads that may be running concurrently in an NVLink cluster. Future iterations of this work should optimize expert and KV tensor placements to minimize travel distance and synchronization overhead.

A natural extension of Harvest is to treat cluster GPU memory as a NUMA-like, non-uniform shared pool, where each device exposes a portion of its HBM as a remotely addressable cache tier. In this model, the key research problem shifts from “offload vs not” to placement and migration under heterogeneous access costs (local HBM, peer HBM over NVLink, host DRAM over PCIe, and potentially CXL-attached memory). The goal is not to eliminate overhead entirely, but to make data movement policy-driven and overlap-friendly, so serving systems can approach the simplicity of a unified pool while retaining performance isolation and correctness under volatility. Promising directions include topology-aware placement, contention-aware bandwidth budgeting, and graceful degradation when remote memory becomes unavailable.

\section{Conclusion}
Harvest uses spare HBM on NVLink-connected peer GPUs as a best-effort cache tier for LLM inference and reduces dependence on PCIe host offload.
Harvest keeps correctness independent of remote placement by pushing durability to the application: an object either has an authoritative host copy or is treated as lossy and reconstructible. Across MoE and KV cache offloading, peer-memory caching reduces transfer latency (up to almost $10\times$ for MoE) and increases throughput by $1.5$--$2.0\times$ for models such as Qwen2-MoE and Phi-3.5-MoE, showing that opportunistic peer HBM can translate memory slack into practical inference speedups.




\section*{Impact Statement}

This paper presents work whose goal is to advance the field of Machine
Learning. There are many potential societal consequences of our work, none of
which we feel must be specifically highlighted here.

The authors declare no ethical considerations or conflicts of interest.

\bibliography{references}
\bibliographystyle{icml2026}

\end{document}